\documentclass[letterpaper]{article} 
\usepackage{aaai2026}  
\usepackage{times}  
\usepackage{helvet}  
\usepackage{courier}  
\usepackage[hyphens]{url}  
\usepackage{graphicx} 
\usepackage{natbib}  
\usepackage{caption} 
\usepackage{algorithm}
\usepackage{algorithmic}
\usepackage{booktabs}
\usepackage{cite}
\usepackage{multirow}
\usepackage{colortbl}
\usepackage{tabularx}
\usepackage{color}
\usepackage[accsupp]{axessibility}  
\usepackage{pifont}
\usepackage{makecell}
\usepackage{lipsum}
\usepackage{threeparttable}
\usepackage{placeins}
\usepackage[svgnames,table]{xcolor}
\usepackage{xspace}
\usepackage{subcaption}
\usepackage{newfloat}
\usepackage{listings}
\DeclareCaptionStyle{ruled}{labelfont=normalfont,labelsep=colon,strut=off} 
\floatstyle{ruled}
\newfloat{listing}{tb}{lst}{}
\floatname{listing}{Listing}
\nocopyright 

\title{RSVG-ZeroOV: Exploring a Training-Free Framework for Zero-Shot Open-Vocabulary Visual Grounding in Remote Sensing Images}
\author{
    Ke Li\textsuperscript{\rm 1},
    Di Wang\textsuperscript{\rm 1}\thanks{Corresponding author.},
    Ting Wang\textsuperscript{\rm 1},
    Fuyu Dong\textsuperscript{\rm 1},
    Yiming Zhang\textsuperscript{\rm 2},
    Luyao Zhang\textsuperscript{\rm 1},
    Xiangyu Wang\textsuperscript{\rm 1},
    Shaofeng Li\textsuperscript{\rm 1},
    Quan Wang\textsuperscript{\rm 1}
}
\affiliations{
    \textsuperscript{\rm 1}School of Computer Science and Technology, Xidian University, 710126, China.\\
    \textsuperscript{\rm 2}Department of Mathematics, University of California, San Diego, USA.
}

\usepackage{bibentry}

\begin{document}

\maketitle

\begin{abstract}
    Remote sensing visual grounding (RSVG) aims to localize objects in remote sensing images based on free-form natural language expressions.
    Existing approaches are typically constrained to closed-set vocabularies, limiting their applicability in open-world scenarios. 
    While recent attempts to leverage generic foundation models for open-vocabulary RSVG, they overly rely on expensive high-quality datasets and time-consuming fine-tuning.
    To address these limitations, we propose \textbf{RSVG-ZeroOV}, a training-free framework that aims to explore the potential of frozen generic foundation models for zero-shot open-vocabulary RSVG.
    Specifically, RSVG-ZeroOV comprises three key stages:
    \textit{(i) Overview:} We utilize a vision-language model (VLM) to obtain cross-attention\footnote[1]{In this paper, although decoder-only VLMs use self-attention over all tokens, we refer to the image-text interaction part as cross-attention to distinguish it from pure visual self-attention.}maps that capture semantic correlations between text queries and visual regions.
    \textit{(ii) Focus:} By leveraging the fine-grained modeling priors of a diffusion model (DM), we fill in gaps in structural and shape information of objects, which are often overlooked by VLM.
    \textit{(iii) Evolve:} A simple yet effective attention evolution module is introduced to suppress irrelevant activations, yielding purified segmentation masks over the referred objects.
    Without cumbersome task-specific training, RSVG-ZeroOV offers an efficient and scalable solution.  
    Extensive experiments demonstrate that the proposed framework consistently outperforms existing weakly-supervised and zero-shot methods.
\end{abstract}

\section{Introduction}
\label{sec:intro}
Deep learning has been widely applied in remote sensing tasks, significantly improving the performance of applications such as image classification, object detection, and semantic segmentation \cite{chen2024rsprompter,li2024show,shan2025ros,pan2025locate}. 
However, these advances are heavily dependent on large-scale annotated data to learn how to recognize and localize objects.
Given the broad geographic coverage and the diverse types of objects present in remote sensing imagery, existing datasets cannot fully capture the diversity of real-world objects. 
As a result, when encountering out-of-distribution objects, existing models may struggle to recognize and localize these objects, limiting their reliability in practical remote sensing scenarios.

\begin{figure}[!t]
    \centering
    \includegraphics[width=\linewidth]{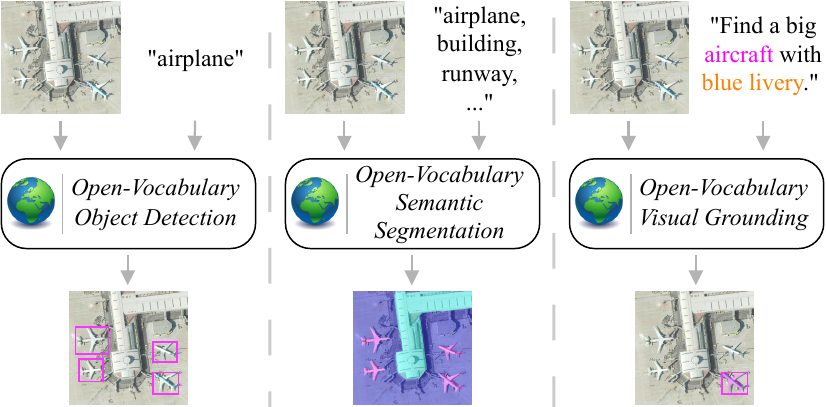}
    \caption{Illustration of different open-vocabulary remote sensing tasks.
    Left and middle panels depict \textit{category-driven} tasks, where models rely on predefined category names to distinguish different image regions.
    Right panel represents \textit{intention-driven} paradigm, enabling users to flexibly specify objects using natural language expressions.}
    \label{fig:teaser}
\end{figure}

Recently, generic foundation models have demonstrated remarkable open-world perception capabilities \cite{liu2023visual,kirillov2023segment,liu2024grounding}. 
Benefiting from pre-training on extensive image-text pairs, these models are able to effectively understand the complex semantic relationships between textual descriptions and visual elements, thereby achieving impressive vision-language alignment.

Some researchers \cite{yao2025remotesam,huang2025zori} have begun exploring the potential of foundation models in open-vocabulary category-driven tasks, such as remote sensing object detection and segmentation, relying on predefined object categories (\textit{e.g.}, “road”, “farmland”) to calculate the similarity with image regions, as illustrated in Fig.~\ref{fig:teaser} (left and middle).
Furthermore, in many real-world scenarios, users are more interested in locating specific targets. For instance, in urban analysis, planners may seek to identify “the tallest building by the river” or “the factory next to the playground.” 
Moreover, some objects cannot be represented by a simple category name. 
For example, temporary roadside parking zones may visually resemble roads but functionally serve as informal parking lots. Such expressions require models to reason about visual attributes, spatial relationships, and functional roles. 
Conventional open-vocabulary detection or segmentation methods are insufficient for handling such intention-driven tasks. 
In contrast, zero-shot open-vocabulary remote sensing visual grounding (RSVG) offers a more practical and flexible solution.

In this work, we focus on a central question: \textit{how to extend the capabilities of frozen foundation models to zero-shot open-vocabulary RSVG?}
To answer this, we conduct a series of exploratory experiments to progressively analyze the impact of different foundation models and architectural components on performance.
Through these explorations, we summarize three empirical guidelines: 
\textbf{1)} Generic VLMs exhibit strong generalization ability, which is critical for object localization in remote sensing. 
\textbf{2)} Compared to other visual models, DMs demonstrate superior perception of the inherent structural concepts of objects. 
\textbf{3)} Attention maps from VLMs and DMs are complementary, and their integration consistently improves grounding performance.

Based on the above guidelines, we propose a training-free framework named \textbf{RSVG-ZeroOV}, a straightforward yet effective approach to solve zero-shot open-vocabulary RSVG. 
Our framework follows an overview-focus-evolve strategy that leverages the distinct yet complementary attention patterns of frozen VLM and DM to form the basis of our grounding results.
Specifically, we first aggregate cross-attention maps from all transformer heads in VLM to highlight regions corresponding to the referred objects.  
However, we observe two major limitations in these attention maps: 
one is that attention tends to concentrate on object boundaries and corners rather than the full extent, while another is that attention is rather scattered, often including irrelevant regions.
To address the first issue, we devise an attention interaction module to facilitate interaction between the cross-attention maps and the self-attention maps, thereby enhancing the activation of pixels corresponding to the referred objects.
For the latter, we propose a simple yet effective attention evolution module to suppress irrelevant activations, resulting in accurate and purified segmentation masks.
Moreover, we incorporate the segment anything model (SAM) as an optional post-processing step to further improve mask quality.
Extensive experiments conducted on two RSVG benchmarks show that RSVG-ZeroOV achieves exceptional performance in open-vocabulary RSVG under zero-shot setting.
\begin{itemize}
  \item We propose RSVG-ZeroOV, a training-free framework for zero-shot open-vocabulary RSVG. We take the pioneering step by integrating attention maps from VLM and DM to enhance the perception of referred objects.
  \item We introduce an overview-focus-evolve strategy to progressively refine the attention maps, enabling the model to effectively capture the referred objects in remote sensing images.
  \item Extensive experiments on two RSVG datasets demonstrate that RSVG-ZeroOV achieves superior zero-shot performance, confirming the effectiveness of foundation models in open-world remote sensing perception.
\end{itemize}

\section{Related work}
\label{sec2}
\textbf{Generic Foundation Model.}
The emergence of large-scale pretraining has fueled the rapid advancement of multimodal foundation models. 
Among them, visual-language models (VLMs) are designed for cross-modal understanding by aligning visual and textual representations. 
Early attempts like CLIP~\cite{radford2021learning} demonstrated impressive zero-shot generalization by contrastively pretraining on massive image-text pairs. 
LLaVA~\cite{liu2023visual} builds upon CLIP by adding a language decoder Vicuna~\cite{chiang2023vicuna} to support multimodal instruction-following.
More recent approaches, such as CogVLM~\cite{wang2024cogvlm}, InternVL~\cite{chen2024internvl}, and Qwen-VL~\cite{wang2024qwen2}, further enhance this paradigm and demonstrate outstanding performance in open-world visual understanding.
However, due to their emphasis on high-level semantic reasoning and lack of pixel-level supervision, these models often struggle with fine-grained spatial perception.
In contrast, another branch represented by diffusion models (DMs) \cite{rombach2022high}, focuses on visual content generation from text prompts.
Recent studies~\cite{pnvr2023ld,xu2023open} reveal that DMs' hierarchical attention encodes strong structural priors, making their attention maps applicable to downstream tasks like classification, detection, and segmentation.
In this study, we conduct systematic exploratory experiments to investigate the potential of integrating VLM and DM models for open-vocabulary RSVG, which will be elaborated in the next section.

\begin{figure*}[!t]
    \centering
    \includegraphics[width=\linewidth]{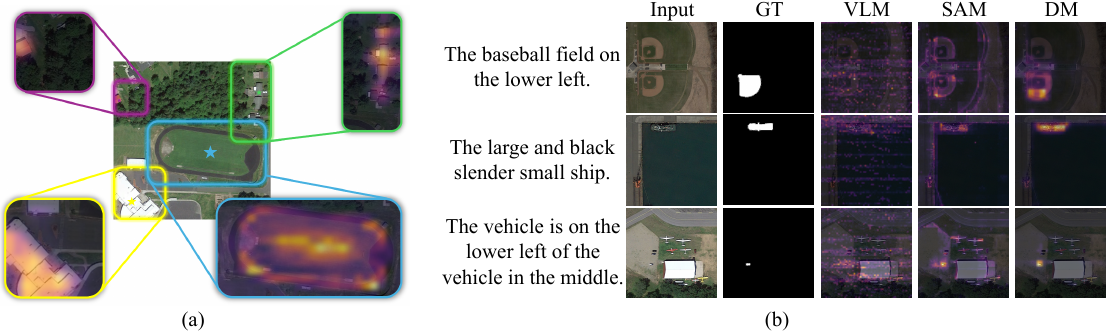}
    \caption{(a) Visualization of receptive fields derived from self-attention maps in DM.  
    (b) Comparison of attention embedding results using self-attention maps from different models.}
    \label{fig:guide2}
\end{figure*}

\noindent \textbf{Remote Sensing Visual Grounding.} 
RSVG aims to localize objects in remote sensing imagery based on natural language descriptions. Compared to natural scene images, remote sensing data is acquired via satellites and often encounters unique challenges such as scale variation and cluttered backgrounds.
Remote sensing referring expression comprehension (RSREC) and segmentation (RSRES) are two typical RSVG tasks. RSREC aims to predict a bounding box corresponding to the given referring expression.
The pioneering work~\cite{sun2022visual} explores RSREC through a one-stage method based on dense~\cite{ding2024visual}. 
Recently, transformer-based methods~\cite{zhan2023rsvg,li2024language,lan2024language,ding2024visual} can effectively capture cross-modal context and achieve better performance, benefiting from the attention mechanism. 
For RSRES, the goal is to predict a precise pixel-wise binary mask specified by the given referring expression. 
Recent works~\cite{yuan2023rrsis,liu2024rotated,dong2024cross,li2025segearth_r1} improve segmentation by enhancing multi-scale visual-text feature interactions.

\noindent \textbf{Open-Vocabulary Visual Perception.} 
With the rapid development of multimodal foundation models, open-vocabulary learning has achieved remarkable success in natural image domains. 
Inspired by this progress, recent studies have extended these paradigms to the remote sensing field.
For instance, RemoteCLIP \cite{liu2024remoteclip} improves transferability in downstream tasks by pretraining on extensive remote sensing image-expression pairs. 
Similarly, VisGT~\cite{pan2025locate} and GSNet~\cite{ye2025towards} construct large-scale open-vocabulary detection and segmentation datasets, respectively, enabling models to recognize a broader range of object categories.
However, curating high-quality annotated datasets for remote sensing remains costly and labor-intensive.
To alleviate reliance on manual annotations, several efforts~\cite{zheng2025instructsam} have explored training-free approaches.
For example, SegEarth-OV~\cite{li2025segearth_ov} adapts existing generic models by modifying or removing task-specific components, allowing seamless application to high-resolution remote sensing segmentation.
Nevertheless, these methods still rely on predefined category labels during inference and lack the flexibility to understand user-intended objects described in free-form language.
As discussed in the introduction, users in practical applications often prefer to locate objects by describing visual attributes, spatial relationships, or functional roles using natural language expressions.
To address this limitation, we introduce the task of zero-shot open-vocabulary RSVG, which requires models to flexibly interpret free-form expressions and localize the referred objects. 
To this end, we propose a training-free framework that integrates attention maps from frozen VLM and DM to facilitate this task.

\section{Guidelines of Exploring Foundation Models}
Naively transferring generic foundation models to open-vocabulary remote sensing visual perception tasks does not yield satisfactory results.  
In this section, we summarize three guidelines for the effective use of foundation models.

\begin{table}[!t]
\centering
\setlength{\tabcolsep}{1.8mm}
{\fontsize{9pt}{9.5pt}\selectfont
\begin{tabular}{l|cc|cc}
    \specialrule{.1em}{.3em}{.3em} 
    \multicolumn{1}{c|}{}                          & \multicolumn{2}{c|}{RSREC} & \multicolumn{2}{c}{RSRES}\\
    \cmidrule(r){2-5}       
    \multicolumn{1}{l|}{Method (Zero-Shot)}        & Pr@0.5 & mIoU & Pr@0.5 & mIoU\\
    \specialrule{.1em}{.3em}{.3em} 
    GeoChat     (CVPR'24)                          & 23.93 & 30.12 & -     & - \\
    LLaVA-1.5   (CVPR'24)                          & 13.21 & 20.59 & -     & - \\
    Qwen2.5-VL    (arXiv'25)                         & 28.66 & 30.90 & -     & - \\
    \midrule
    GeoChat + SAM                                  & 27.61 & 32.53 & 18.85 & 24.92 \\
    LLaVA-1.5 + SAM                                & 16.26 & 21.97 & 10.86 & 16.86 \\
    Qwen2.5-VL + SAM                                 & 30.35 & 31.93 & 24.68 & 25.72 \\
    \specialrule{.1em}{.3em}{.3em} 
\end{tabular}}
\caption{Comparison of generic VLMs and remote sensing VLM on RRSIS-D test set, all using 7B-scale LLMs.}
\label{tab:guide1}
\end{table}

\noindent \textbf{Guideline 1: generic VLMs exhibit strong generalization ability on unseen remote sensing data.}
To demonstrate this, we conduct zero-shot experiments on RRSIS-D test set. 
Specifically, we compare three 7B-scale models, including LLaVA-1.5~\cite{liu2024improved}, Qwen2.5-VL~\cite{Qwen2.5-VL}, and GeoChat~\cite{kuckreja2024geochat}. 
For RSREC task, we generate bounding box predictions of referred objects using a unified prompt format: [Locate the object referred to by ‘\{referring expression\}’ and return its box coordinates (x1, y1, x2, y2)].
Notably, GeoChat has been finetuned on 318k image-instruction pairs collected from remote sensing datasets\footnote[2]{A portion of GeoChat's training data shares source images (from DIOR~\cite{li2020object} dataset) with RRSIS-D test set, making the evaluation not strictly zero-shot for GeoChat.}, enabling it to learn more domain-specific priors.
However, as shown in Tab.~\ref{tab:guide1}, GeoChat achieves 23.93\% Pr@0.5 on RSREC task, noticeably lower than Qwen2.5-VL's \textbf{28.66\%}. 
This result suggests that VLMs pretrained on large-scale generic datasets can exhibit strong cross-domain generalization even without access to remote sensing data.
For RSRES task, we utilize outputs of VLMs as box prompts for SAM to get pixel-level masks.
As we can see, Qwen2.5-VL still outperforms GeoChat by \textbf{0.8\%} in mIoU.
These findings highlight the potential of generic VLMs as a solid foundation for zero-shot open-vocabulary RSVG, even without any domain-specific training.

\noindent \textbf{Guideline 2: self-attention maps in DM encode superior structural priors of objects.}
Theoretically, self-attention mechanisms are inherently suited for modeling object structure, as they compute pairwise similarities across visual elements, enabling the network to capture spatial dependencies and shape layouts.
As shown in Fig.~\ref{fig:guide2}(a), we visualize the receptive fields of self-attention maps anchored at different query points. 
The results demonstrate that self-attention can effectively distinguish foreground objects from the background, reflecting a strong capacity for structural perception.
To further investigate this, we compare the spatial distributions of self-attention extracted from three representative architectures: the visual encoder of VLM, the image backbone of SAM, and the U-Net of DM, as illustrated in Fig.~\ref{fig:guide2}(b). 
Specifically, the VLM attention maps tend to be broadly distributed and spatially scattered, consistent with its objective of capturing global semantic context.
While SAM produces sharper boundaries, its purely visual design lacks semantic understanding, which often leads to over-attention on surrounding background areas.
For instance, in the second row, SAM simultaneously highlights both the ship and port. 
In contrast, the self-attention maps from DM display the most coherent structural representations, \textit{i.e.,} attention is evenly and densely distributed across the entire object extent, revealing a more precise and holistic understanding of object shapes.
To quantitatively validate these observations, we integrate the different types of self-attention maps into our visual grounding pipeline and evaluate their performance.
As reported in Tab.~\ref{tab:guide23}, the use of self-attention maps from DM (the last row) achieves the highest Pr@0.5 and mIoU on both RSREC and RSRES benchmarks.
These results confirm the superiority of DM's self-attention in capturing object-centric structural cues for visual grounding tasks.

\begin{table}[!t]
\centering
\setlength{\tabcolsep}{1.6mm}
{\fontsize{9pt}{9.5pt}\selectfont
\begin{tabular}{l|cc|cc}
    \specialrule{.1em}{.3em}{.3em} 
    \multicolumn{1}{c|}{}                         & \multicolumn{2}{c|}{RSREC} & \multicolumn{2}{c}{RSRES}\\
    \cmidrule(r){2-5}       
    \multicolumn{1}{l|}{Self-Attention Maps} & Pr@0.5 & mIoU & Pr@0.5 & mIoU\\
    \specialrule{.1em}{.3em}{.3em}
    \textit{w/o} Cross-Attention Map                     & 21.49 & 26.26 & 1.18  & 6.15 \\
    ViT in VLM                       & 12.53 & 21.02 & 2.13  & 7.40\\
    ViT in SAM                       & 18.85 & 23.56 & 4.77  & 10.23 \\
    U-Net in DM (Ours)               & \textbf{30.15} & \textbf{32.92} & \textbf{12.84} & \textbf{21.85}   \\
    \specialrule{.1em}{.3em}{.3em} 
\end{tabular}}
\caption{Evaluation of different self-attention maps on RRSIS-D test set.}
\label{tab:guide23}
\end{table}

\noindent \textbf{Guideline 3: combining cross- and self-attention maps is vital especially for pixel-level perception tasks.}
To validate this, we first evaluate the grounding performance using only the cross-attention maps from the VLM, as shown in Tab.~\ref{tab:guide23} (the first row). 
Removing the self-attention maps from DM leads to significant performance drops of \textbf{8.66\%} Pr@0.5 and \textbf{6.66\%} mIoU on RSREC, and \textbf{11.66\%} and \textbf{15.70\%} on RSRES, respectively.
Such a phenomenon indicates the necessity of combining cross- and self-attention maps, especially for pixel-level perception tasks like RSRES.
These findings naturally lead to a question: \textit{which interaction strategy best integrates cross- and self-attention maps for optimal grounding performance?}
We try four strategies: 
1) Anchor-based: selecting high-response pixels from the cross-attention map to represent the object, followed by aggregating their corresponding self-attention regions.
2) Multiplication: performing element-wise multiplication on cross- and self-attention maps, denoted as $\mathcal{A}_S \cdot \mathcal{A}_C$. 
3) Exponentiation: enhancing the self-attention map by raising it to a power $\gamma$, \textit{i.e.}, $\mathcal{A}_S^{\gamma} \cdot \mathcal{A}_C$. 
4) Similarity: computing cosine similarity between the cross- and self-attention maps at each spatial location (details in the next section).
As summarized in Tab.~\ref{tab:guide3}, although the anchor-based and exponentiation strategies achieve reasonable performance on RSREC, they suffer on RSRES. The former tends to oversimplify by omitting many regions, while the latter tends to overamplify high activations. 
In contrast, the proposed similarity strategy yields more semantically consistent initial masks, which allows the evolve stage to further improve segmentation quality, ultimately achieving the best performance on both benchmarks.

\begin{table}[!t]
\centering
\setlength{\tabcolsep}{2.2mm}
{\fontsize{9pt}{9.5pt}\selectfont
\begin{tabular}{l|cc|cc}
    \specialrule{.1em}{.3em}{.3em} 
    \multicolumn{1}{c|}{}                         & \multicolumn{2}{c|}{RSREC} & \multicolumn{2}{c}{RSRES}\\
    \cmidrule(r){2-5}       
    \multicolumn{1}{l|}{Interaction Strategy} & Pr@0.5 & mIoU & Pr@0.5 & mIoU\\
    \specialrule{.1em}{.3em}{.3em}
    Anchor-based                        & 23.18 & 26.94 & 6.58  & 16.69 \\
    \ \ + \textit{Evolve stage}         & 28.73 & 31.48 & 7.61  & 16.58 \\
    \midrule
    Multiplication                      & 21.94 & 26.32 & 10.03 & 19.78 \\
    \ \ + \textit{Evolve stage}         & 29.26 & 31.54 & 11.52 & 20.75 \\
    \midrule
    Exponentiation                      & 24.07 & 28.61 & 5.37  & 15.30 \\
    \ \ + \textit{Evolve stage}         & 27.38 & 31.02 & 5.31  & 14.00 \\
    \midrule
    \textit{Similarity} (Ours)          & 22.63 & 26.65 & 10.26 & 20.56 \\
    \ \ + \textit{Evolve stage}         & \textbf{30.15} & \textbf{32.92} & \textbf{12.84} & \textbf{21.85}   \\
    \specialrule{.1em}{.3em}{.3em} 
\end{tabular}}
\caption{Comparison of different interaction strategies between cross- and self-attention maps on RRSIS-D test set.}
\label{tab:guide3}
\end{table}

\begin{figure*}[!t]
    \centering
    \includegraphics[width=\linewidth]{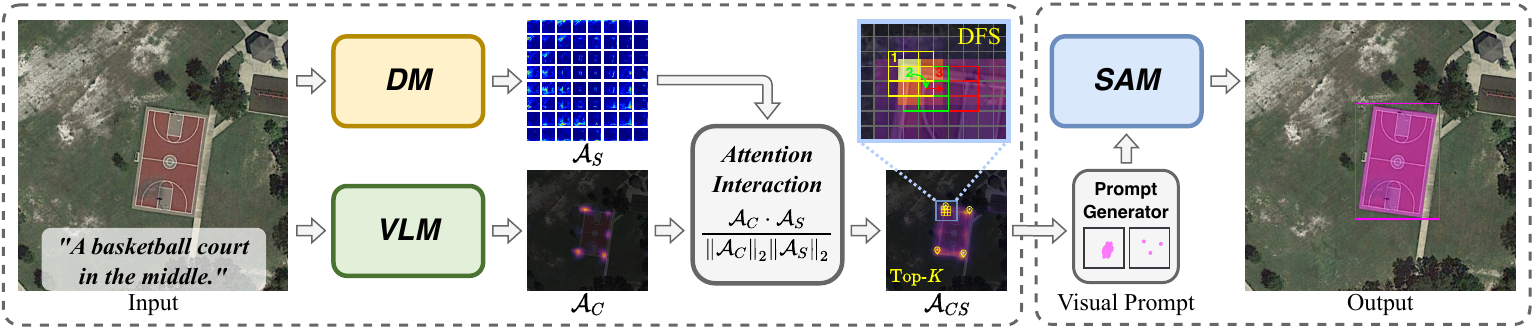}
    \caption{The framework of the proposed RSVG-ZeroOV.} 
    \label{fig:framework}
\end{figure*}

\section{Proposed Framework}
Following our proposed guidelines, we present a straightforward training-free framework named \textbf{RSVG-ZeroOV} for zero-shot open-vocabulary RSVG.
As illustrated in Fig.~\ref{fig:framework}, the framework comprises three sequential stages:
\textit{(i) Overview:} we prompt a VLM with a referring expression to identify the referred object and extract its corresponding cross-attention map, which highlights regions most relevant to the given textual query.
\textit{(ii) Focus:} we cache the self-attention maps from the U-Net of a DM to obtain structural priors of objects, and then propose an effective interaction strategy to embed these priors into the cross-attention distribution.
\textit{(iii) Evolve:} to generate a high-quality attention map, we introduce an attention evolution module to filter out irrelevant activations, thus improving the quality of the segmentation mask.

\begin{figure}[!t]
    \centering
    \includegraphics[width=\linewidth]{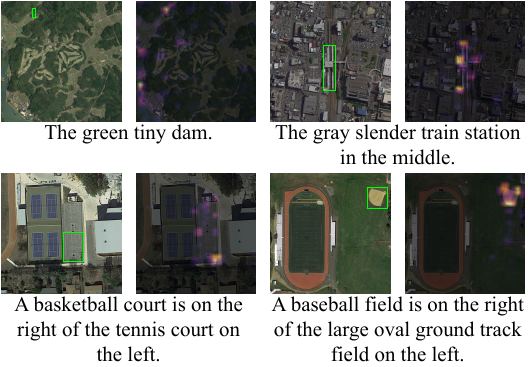}
    \caption{Visualization of some cross-attention maps from VLM on RRSIS-D test set.}
    \label{fig:guide1weak}
\end{figure}

\textbf{Overview Stage.} 
Referring to \textbf{Guideline 1}, we utilize the attention map from a frozen VLM to capture the coarse-level localization cues of the referred object.
Given a remote sensing image $\mathcal{I}$ and a referring expression $\mathcal{T}$, we first hook into the attention weights $\mathcal{W}^{(t)} \in \mathbb{R}^{H \times 1 \times N}$ from VLM, where $N$ is the token length, $H$ is the number of attention heads, and $t \in \{1, \dots, T\}$ indexes the $T$ autoregressive forward passes.
To obtain cross-attention parts, we extract the image-text-related segment $\mathcal{W}^{(t)}_{p:p'}$, where $p$ and $p'$ indicate the span of visual tokens in the input sequence.
After that, we aggregate information from the transformer heads with mean weights, and form a sentence-level cross-attention map by averaging all $t$ as:
\begin{equation}
  \label{eq:equ1}
  \mathcal{A}_C = 1/T {\textstyle \sum_{t=1}^{T}} ( 1/H {\textstyle \sum_{h=1}^{H}} \mathcal{W}^{(t)}_{p:p'} ).
\end{equation}
As shown in Fig.~\ref{fig:guide1weak}, cross-attention maps establish the relationship between textual queries and visual pixels. 
However, it exhibits two key limitations:
\textit{(i) Attention tends to focus on object boundaries or corners rather than the full extent of the object.} 
This behavior stems from the high-level semantic concentration of VLMs, which encourages attention to concentrate on key features of objects.
\textit{(ii) Attention is often scattered and includes irrelevant regions.} 
This dispersion arises because VLMs need to aggregate contextual cues from multiple visual regions to understand complex referring expressions.

\textbf{Focus Stage.} 
To mitigate the first \textit{issue (i)}, referring to the aforementioned \textbf{Guidelines 2\&3}, we propose an attention interaction module that integrates cross-attention with structural priors derived from self-attention. 
Considering the diverse scales of objects in remote sensing images, we first extract multi-scale self-attention maps from the U-Net in DM, and fuse them to form a unified structural prior $\mathcal{A}_S\in \mathbb{R}^{H \times W \times H \times W}$, which is denoted as:
\begin{equation}
\mathcal{A}_S = 1/L {\textstyle \sum_{l\in L}}(\mathcal{A}^{l}_S),
\end{equation}
where $\mathcal{A}_S^l$ denotes the self-attention map from layer $l$.
Next, we calculate the correlation score\footnote[3]{In fact, $\mathcal{A}_C$ does not match the spatial resolution of $\mathcal{A}_S$, hence we interpolate $\mathcal{A}_C$ to the same size as $\mathcal{A}_S$ before interaction.} between $\mathcal{A}_C$ and $\mathcal{A}_S$ through cosine similarity:
\begin{equation}
  \label{eq:equ2}
  \mathcal{A}_{CS} = \frac{\mathcal{A}_C \cdot \mathcal{A}_S}{\|\mathcal{A}_C\|_2 \|\mathcal{A}_S\|_2},
\end{equation}
where $\|\cdot\|_2$ denotes the $L_2$ norm. 
This interaction introduces object-centric structural guidance into the original cross-attention map, allowing the model to shift focus from incomplete boundaries to semantically coherent regions.

\begin{table*}[!t]
\centering
\setlength{\tabcolsep}{1.8mm}
{\fontsize{9pt}{9.5pt}\selectfont
\begin{tabular}{c|l|ccccc|ccccc}
    \specialrule{.1em}{.3em}{.3em} 
    \multicolumn{2}{c|}{}      & \multicolumn{5}{c|}{RSREC}    & \multicolumn{5}{c}{RSRES}\\
    \cmidrule(r){3-12}       
    \multicolumn{2}{l|}{Method}                        & Pr@0.3& Pr@0.5& Pr@0.7& mIoU  & oIoU  &Pr@0.3 & Pr@0.5 & Pr@0.7& mIoU  & oIoU\\
    \specialrule{.1em}{.3em}{.3em} 
    \multirow{3}{*}{\textit{W-S}} 
    & TRIS         (ICCV'23)                           & 14.62 & 3.79  & 0.37  & 13.20 & 13.14 & 15.02 & 4.91   & 1.26  & 13.11 & 15.40   \\ 
    & SAG          (ICCV'23)                           & 9.39  & 2.10  & 0.29  & 9.22  & 9.21  & 11.63 & 4.02   & 0.60  & 11.10 & 11.13    \\ 
    & QueryMatch   (MM'24)                             & 22.04 & 16.22 & 12.10 & 17.21 & 15.26 & 20.97 & 15.54  & 10.62 & 15.73 & 10.73    \\ 
    \midrule 
    \multirow{7}{*}{\textit{Z-S}}  
    & VLM (Baseline)                                   & 43.35 & 28.66 & 14.27 & 30.90 & 29.53 & -     & -      & -     & -     & -      \\ 
    & DiﬀSegmenter (TIP'25)                            & 8.22  & 1.69  & 0.20  & 8.70  & 8.63  & 2.18  & 0.23   & 0.00  & 4.47  & 4.35    \\ 
    & \textcolor{gray}{DiﬀSegmenter (\textit{w/ VLM})} & \textcolor{gray}{39.80} & \textcolor{gray}{24.29} & \textcolor{gray}{13.01} & \textcolor{gray}{28.47} & \textcolor{gray}{26.92} & \textcolor{gray}{24.99} & \textcolor{gray}{9.48} & \textcolor{gray}{1.61} & \textcolor{gray}{19.03} & \textcolor{gray}{15.19}\\
    & DiffPNG      (ECCV'24)                           & 18.93 & 8.45  & 3.30  & 14.84 & 12.54 & 17.70 & 8.27   & 3.02  & 14.86 & 14.71   \\ 
    & \textcolor{gray}{DiffPNG (\textit{w/ VLM})}      & \textcolor{gray}{37.92} & \textcolor{gray}{23.18} & \textcolor{gray}{11.35} & \textcolor{gray}{26.93} & \textcolor{gray}{25.65} & \textcolor{gray}{19.25} & \textcolor{gray}{6.78}   & \textcolor{gray}{0.95}  & \textcolor{gray}{15.49} & \textcolor{gray}{10.55}   \\
    & OV-VG (arXiv'24)                                 & 27.78 & 19.16 & 12.96 & 22.07 & 18.34 & -     & -      & -     & -     & -      \\
    & GroundVLP (AAAI'24)                              & 21.66 & 16.83 & 12.27 & 17.14 & 17.51 & -     & -      & -     & -     & -      \\
    \rowcolor{Lavender} \cellcolor{white} &  \textbf{RSVG-ZeroOV (Ours)}       & \textbf{45.71} & 30.15 & 16.74 & \underline{32.92} & \textbf{32.94} & 30.97 & 12.84  & 2.36  & 21.85 & 18.85   \\  
    \midrule 
    \multirow{6}{*}{\textit{Z-S}}  
    & VLM (Baseline)                                   & 43.60 & \underline{30.35} & \underline{16.76} & 31.93 & 28.38 & \underline{38.18} & \underline{24.68}  & \underline{11.00} & \underline{25.72} & \underline{20.46}   \\ 
    & DiﬀSegmenter (TIP'25)                            & 8.16  & 1.55  & 0.17  & 8.60  & 8.58  & 3.02  & 0.32   & 0.03  & 4.60  & 4.70\\ 
    & \textcolor{gray}{DiﬀSegmenter (\textit{w/ VLM})} & \textcolor{gray}{38.41} & \textcolor{gray}{25.11} & \textcolor{gray}{13.96} & \textcolor{gray}{28.50} & \textcolor{gray}{24.57} & \textcolor{gray}{32.92} & \textcolor{gray}{19.42} & \textcolor{gray}{9.14} & \textcolor{gray}{23.73} & \textcolor{gray}{18.00}\\
    \multirow{4}{*}{\textit{w/ Refine}}  
    & DiffPNG      (ECCV'24)          & 19.59 & 10.31  & 5.03  & 15.71 & 13.11 & 19.65 & 11.32 & 6.23  & 16.22 & 13.77   \\ 
    & \textcolor{gray}{DiffPNG (\textit{w/ VLM})}      & \textcolor{gray}{32.43}    & \textcolor{gray}{21.29}  & \textcolor{gray}{13.04} & \textcolor{gray}{24.89} & \textcolor{gray}{20.79} & \textcolor{gray}{27.03} & \textcolor{gray}{17.64}  & \textcolor{gray}{\underline{11.00}} & \textcolor{gray}{20.99} & \textcolor{gray}{13.53}   \\
    & OV-VG (arXiv'24)                                 & 27.23 & 16.20 & 13.27 & 21.62 & 17.18 & 20.68 & 15.51 & 9.05  & 16.17 & 9.71  \\
    & GroundVLP (AAAI'24)                              & 21.26 & 16.20 & 11.32 & 16.51 & 16.88 & 18.07 & 13.33 & 7.24  & 13.14 & 11.23 \\
    \rowcolor{Lavender} \cellcolor{white} &  \textbf{RSVG-ZeroOV (Ours)}       & \underline{45.70} & \textbf{31.39}  & \textbf{17.63} & \textbf{34.49} & \underline{31.28} & \textbf{40.01} & \textbf{27.39}  & \textbf{13.38} & \textbf{28.35} & \textbf{22.83}  \\ 
    \specialrule{.1em}{.3em}{.3em} 
\end{tabular}}
\caption{Comparisons with SOTA methods on \textbf{RRSIS-D} test set. 
‘\textit{W-S}’ and ‘\textit{Z-S}’ denote the weakly-supervised and zero-shot settings, respectively.
‘\textit{w/ VLM}’ indicates that the cross-attention from the VLM is used to replace the original cross-attention.
‘\textit{w/ Refine}’ indicates that post-processing operations are applied to further refine the segmentation results.}
\label{tab:RRSIS-D}
\end{table*}

\textbf{Evolve Stage.} 
Although this interaction strategy remarkably fills the gaps within the object area, it also exacerbates the impact of \textit{issue (ii)}.
To address this, we introduce an attention evolution module based on recursive region expansion to suppress scattered activation signals that are outside the referred object area.
Specifically, we first select the top-$K$ pixels from the interaction map $\mathcal{A}_{C}$ with the highest response values:
\begin{equation}
    \label{eq:equ3}
    \mathcal{S} = \mathrm{TopK}(\mathcal{A}_{C}, K),
\end{equation}
where $\mathcal{S} = \{(i_k, j_k)\}_{k=1}^{K}$ denotes the set of starting seed positions.
For each seed $(i_k, j_k)$, we define a local neighborhood kernel $\mathcal{N}(i, j)$ of size 3$\times$3, and iteratively expand the region by selecting the neighbor with the highest response:
\begin{equation}
    \label{eq:equ4}
    (i^*, j^*) =\underset{(u,v) \in \mathcal{N}(i,j)}{\arg\max} \mathcal{A}_{CS}[u, v].
\end{equation}
We then perform a depth-first search (DFS) starting from each seed to recursively grow the region. 
A pixel $(u,v)$ is included in the evolved attention map $\mathcal{A}_E$ if it satisfies:
\begin{equation}
\mathcal{A}_{CS}[u, v] \geq \tau,\ \text{\&\&} \ (u,v) \in \mathrm{DFS}(\mathcal{S}),
\end{equation}
where $\tau$ is a response threshold that controls the expansion boundary, and the process continues until no further qualified neighbors can be added.
In addition, we conduct an ablation study to analyze the impact of the ordering between the \textbf{Focus} and \textbf{Evolve} stages on performance (see Tab.~\ref{tab:ablation1} for details).
Finally, we perform binarization on the evolved attention map $\mathcal{A}_E$ to obtain the segmentation mask $\mathbf{M} \in \{0, 1\}^{H \times W}$, which is defined as:
\begin{equation}
    \label{eq:equ8}
    \mathbf{M}(i, j) = 
    \begin{cases}
    1, & \text{if } \mathcal{A}_E(i, j) > \alpha \\
    0, & \text{otherwise}
    \end{cases},
\end{equation}
where $\alpha \in [0, 1]$ is a predefined binarization threshold.

\textbf{Refine Stage (Optional).} 
Following prior work~\cite{liu2024rotated,dong2024cross}, we additionally introduce an optional refinement stage to further improve the segmentation mask $\mathbf{M}$.
We evaluate the effectiveness of several refinement methods, including DenseCRF and various prompt types for SAM.
Our results show that box prompts yield the best performance, whereas point prompts are less effective.

\begin{table}[!t]
\centering
\setlength{\tabcolsep}{0.8mm}
{\fontsize{9pt}{9.5pt}\selectfont
\begin{tabular}{l|cc|cc}
    \specialrule{.1em}{.3em}{.3em} 
    \multicolumn{1}{c|}{}                         & \multicolumn{2}{c|}{RSREC} & \multicolumn{2}{c}{RSRES}\\
    \cmidrule(r){2-5}       
    \multicolumn{1}{l|}{Method} & Pr@0.5 & mIoU & Pr@0.5 & mIoU\\
    \specialrule{.1em}{.3em}{.3em} 
    GeoChat     (CVPR'24) + SAM          & 27.61 & 32.53 & 18.85 & 24.92 \\
    Qwen2.5-VL   (arXiv'25) + SAM          & 30.35 & 31.93 & 24.68 & 20.46 \\
    LISA        (CVPR'24)          & 25.80 & 27.78 & 24.51 & 26.78 \\
    NExT-Chat   (ICML'24)          & -     & -     & 26.37 & 24.98 \\
    \rowcolor{Lavender} \textbf{RSVG-ZeroOV (Ours)}        & \textbf{31.39} & \textbf{34.49} & \textbf{27.39} & \textbf{28.35} \\ 
    \specialrule{.1em}{.3em}{.3em} 
\end{tabular}}
\caption{Comparisons with SOTA generic pixel-level VLMs and a remote sensing VLM on \textbf{RRSIS-D} test set.}
\label{tab:appendixresults}
\end{table}

\begin{table*}[!t]
\centering
\setlength{\tabcolsep}{1.5mm}
{\fontsize{9pt}{9.5pt}\selectfont
\begin{tabular}{c|l|ccccc|ccccc}
    \specialrule{.1em}{.3em}{.3em} 
    \multicolumn{2}{c|}{}      & \multicolumn{5}{c|}{RSREC}    & \multicolumn{5}{c}{RSRES}\\
    \cmidrule(r){3-12}       
    \multicolumn{2}{l|}{Method}                             & Pr@0.3& Pr@0.5 & Pr@0.7& mIoU  & oIoU  &Pr@0.3 & Pr@0.5& Pr@0.7& mIoU  & oIoU\\
    \specialrule{.1em}{.3em}{.3em} 
    \multirow{3}{*}{\textit{W-S}} 
    & TRIS         (ICCV'23)          & 19.27 & 8.89  & 2.86  & 14.63 & 15.74 & 14.23 & 4.40   & 0.92  & 11.46 & 16.06   \\ 
    & SAG          (ICCV'23)          & 5.83  & 1.78  & 0.33  & 7.12  & 7.10  & 8.79  & 2.55   & 0.35  & 9.31  & 9.36    \\ 
    & QueryMatch   (MM'24)            & 31.79 & 27.74 & 23.32 & 26.72 & 15.86 & 31.06 & 26.68  & \textbf{20.27} & 24.59 & 10.77    \\ 
    \midrule 
    \multirow{7}{*}{\textit{Z-S}}  
    & VLM (Baseline)                  & 48.62 & 35.08 & 20.26 & 34.60 & 30.15 & -     & -      & -     & -     & -      \\ 
    & DiﬀSegmenter (TIP'25)           & 6.03  & 1.87  & 0.40  & 7.14  & 7.00  & 2.44  & 0.46  & 0.06  & 4.00  & 3.74    \\ 
    & \textcolor{gray}{DiﬀSegmenter (\textit{w/ VLM})}& \textcolor{gray}{27.17} & \textcolor{gray}{15.93} & \textcolor{gray}{7.98} & \textcolor{gray}{21.61} & \textcolor{gray}{19.03} & \textcolor{gray}{28.65} & \textcolor{gray}{11.04} & \textcolor{gray}{2.50} & \textcolor{gray}{21.37} & \textcolor{gray}{\underline{21.87}} \\
    & DiffPNG      (ECCV'24)          & 12.14 & 6.03  & 2.72  & 11.10 & 9.01  & 13.44 & 5.79  & 1.93  & 11.85 & 11.79   \\ 
    & \textcolor{gray}{DiffPNG (\textit{w/ VLM})}     & \textcolor{gray}{34.60} & \textcolor{gray}{21.54} & \textcolor{gray}{11.62} & \textcolor{gray}{26.18} & \textcolor{gray}{23.58} & \textcolor{gray}{23.50} & \textcolor{gray}{8.73} & \textcolor{gray}{1.76} & \textcolor{gray}{18.49} & \textcolor{gray}{14.45} \\
    & OV-VG (arXiv'24)                                & 26.83 & 22.16 & 18.31 & 22.65 & 15.64 & -     & -      & -     & -     & -      \\
    & GroundVLP (AAAI'24)                             & 23.13 & 19.90 & 17.15 & 19.76 & 16.46 & -     & -      & -     & -     & -      \\
    \rowcolor{Lavender} \cellcolor{white} &  \textbf{RSVG-ZeroOV (Ours)}      & \underline{50.17} & 35.08 & 19.03 & 36.20 & \textbf{34.37} & 32.71 & 14.74  & 3.73  & 22.69 & 21.75   \\  
    \midrule 
    \multirow{6}{*}{\textit{Z-S}}  
    & VLM (Baseline)                  & 49.07 & \underline{38.09} & \textbf{26.57} & \underline{37.74} & 30.68 & \underline{43.03} & \underline{30.38} & 18.04 & \underline{30.45} & 21.30  \\ 
    & DiﬀSegmenter (TIP'25)           & 6.05  & 1.99  & 0.45  & 7.11  & 6.96  & 2.65  & 0.73   & 0.17  & 3.88  & 3.79 \\ 
    & \textcolor{gray}{DiﬀSegmenter (\textit{w/ VLM})}& \textcolor{gray}{26.74} & \textcolor{gray}{16.74} & \textcolor{gray}{9.47} & \textcolor{gray}{22.01} & \textcolor{gray}{17.91} & \textcolor{gray}{21.39} & \textcolor{gray}{12.02} & \textcolor{gray}{6.40} & \textcolor{gray}{17.65} & \textcolor{gray}{12.87} \\
    \multirow{4}{*}{\textit{w/ Refine}}  
    & DiffPNG      (ECCV'24)          & 9.34  & 3.76  & 1.38  & 8.43  & 8.66  & 3.99  & 1.73  & 0.76  & 4.53 & 5.81   \\ 
    & \textcolor{gray}{DiffPNG (\textit{w/ VLM})}     & \textcolor{gray}{22.07}    & \textcolor{gray}{12.19}  & \textcolor{gray}{6.00} & \textcolor{gray}{17.16} & \textcolor{gray}{17.40} & \textcolor{gray}{14.74} & \textcolor{gray}{5.65}  & \textcolor{gray}{1.60} & \textcolor{gray}{12.19} & \textcolor{gray}{8.44}   \\
    & OV-VG (arXiv'24)                                & 26.90 & 22.40 & 18.50 & 22.85 & 15.50 & 23.67 & 17.75 & 12.25 & 18.16 & 9.20  \\
    & GroundVLP (AAAI'24)                             & 23.13 & 19.91 & 16.37 & 19.19 & 16.17 & 20.78 & 15.82 & 10.95 & 15.58 & 10.08 \\
    \rowcolor{Lavender} \cellcolor{white} &  \textbf{RSVG-ZeroOV (Ours)}      & \textbf{50.77} & \textbf{38.90}  & \underline{24.93} & \textbf{38.87} & \underline{34.30} & \textbf{44.30} & \textbf{31.03}  & \underline{18.61} & \textbf{31.84} & \textbf{26.35}  \\ 
    \specialrule{.1em}{.3em}{.3em} 
\end{tabular}}
\caption{Comparisons with SOTA methods on \textbf{RISBench} test set. 
‘\textit{W-S}’ and ‘\textit{Z-S}’ denote the weakly-supervised and zero-shot settings, respectively.
‘\textit{w/ VLM}’ indicates that the cross-attention from the VLM is used to replace the original cross-attention.
‘\textit{w/ Refine}’ indicates that post-processing operations are applied to further refine the segmentation results.}
\label{tab:RISBench}
\end{table*}

\section{Experiments}
\label{sec4}
\subsection{Experimental Settings}
\textbf{Datasets.}
We evaluate our method on two benchmarks tailored for RSVG: RRSIS-D \cite{liu2024rotated} and RISBench \cite{dong2024cross}.
Each sample comprises an RGB image, a referring expression, and a segmentation mask. 
Both datasets are extended from the RSREC dataset, inheriting its bounding box annotations, which allows for comprehensive evaluation of both REREC and RSRES tasks.

\noindent \textbf{Implementation Details.}
Our RSVG-ZeroOV offers a zero-shot solution, requiring only an inference process, without the need for any training images or annotations.
We employ the pre-trained Qwen2.5-VL \cite{Qwen2.5-VL} and Stable Diffusion V1.4 \cite{rombach2022high} as our VLM and DM, respectively.
For Stable Diffusion, we use a guidance scale of 7.5 with 1,000 total diffusion steps, and perform DDIM-based sampling with 20 steps. The DDIM noise schedule follows a scaled linear progression from 0.00085 to 0.012.
Following prior works~\cite{li2024language, liu2024rotated}, we report mIoU, oIoU and Precision@X with $X \in \{0.3,0.5,0.7\}$.
We set $K$=7 for seed selection, $\tau$=0.3 for response thresholding, and $\alpha$=0.4 for binarization.
All experiments are conducted on a single RTX4090 GPU.

\subsection{Main Results}
\noindent \textbf{Comparison Results on RRSIS-D.}
As shown in Tab.~\ref{tab:RRSIS-D}, RSVG-ZeroOV (\textit{w/ Refine}) achieves SOTA performance under the zero-shot setting, surpassing all weakly-supervised and zero-shot methods.
Compared to the best weakly-supervised method QueryMatch, our approach achieves substantial gains in mIoU \textbf{+17.28\%} on RSREC and \textbf{+12.62\%} on RSRES, highlighting the strong grounding capability of our zero-shot framework.
Compared with other zero-shot approaches, \textit{i.e.}, DiffSegmenter (\textit{w/ VLM}) and DiffPNG (\textit{w/ VLM}), our method also achieves significant improvements in mIoU by \textbf{4.45\%-5.99\%} on RSREC, and \textbf{2.82\%-6.36\%} on RSRES, respectively.
This highlights the effectiveness of our framework and its related components beyond VLM alone.
Furthermore, we compare our method against several SOTA generic VLMs and a remote sensing VLM, as shown in Tab.~\ref{tab:appendixresults}. RSVG-ZeroOV achieves the best performance across all metrics on both RSREC and RSRES, demonstrating superior grounding ability even compared to models tailored for pixel-level reasoning or remote sensing tasks.

\noindent \textbf{Comparison Results on RISBench.} 
We further evaluate RSVG-ZeroOV on the more challenging RISBench, which features longer and more semantically complex referring expressions compared to RRSIS-D.
As shown in Tab.~\ref{tab:RISBench}, RSVG-ZeroOV (\textit{w/ Refine}) demonstrates exceptional performance, establishing new SOTA benchmarks for Pr@0.5 and mIoU at \textbf{38.90\%/38.87\%} and \textbf{31.03\%/31.84\%} on both tasks, respectively.
Notably, the proposed method outperforms the two zero-shot methods, OV-VG and GroundVLP by \textbf{+18.8\%/18.13\%} and \textbf{+17.15\%/16.27\%} in oIoU, respectively, indicating its superior ability to capture object structures and preserve overall shape integrity.

\begin{table}[!t]
\centering
\setlength{\tabcolsep}{1.8mm}
{\fontsize{9pt}{9.5pt}\selectfont
\begin{tabular}{l|cc|cc}
    \specialrule{.1em}{.3em}{.3em} 
    \multicolumn{1}{c|}{}       & \multicolumn{2}{c|}{RSREC} & \multicolumn{2}{c}{RSRES}\\
    \cmidrule(r){2-5}       
    \multicolumn{1}{l|}{Method} & Pr@0.5 & mIoU & Pr@0.5 & mIoU\\
    \specialrule{.1em}{.3em}{.3em} 
    RSVG-ZeroOV (Ours)       & \textbf{30.15} & \textbf{32.92} & \textbf{12.84} & \textbf{21.85}   \\
    \textit{w/o} VLM         & 16.22 & 18.82 & 11.43 & 15.81 \\
    \textit{w/o} DM          & 21.49 & 26.26 &  1.18 &  6.15 \\
    \textit{w/o} Evolve      & 22.63 & 26.65 & 10.26 & 20.56 \\
    \midrule
    \textit{O-F-E} Stage   & \textbf{30.15} & \textbf{32.92} & \textbf{12.84} & \textbf{21.85}\\
    \textit{O-E-F} Stage   & 27.34 & 29.51 & 7.18 & 15.89\\
    \specialrule{.1em}{.3em}{.3em} 
\end{tabular}}
\caption{Effect of components in RSVG-ZeroOV.}
\label{tab:ablation1}
\end{table}

\subsection{Ablation Study}
\label{sec4c}
In this section, we report ablation study results on the RRSIS-D test set to investigate its effectiveness.

\noindent \textbf{Effect of RSVG-ZeroOV's components.}
As shown in Tab.~\ref{tab:ablation1}, removing VLM (replaced by a cross-attention map in DM) leads to significant drops of \textbf{14.10\%} and \textbf{6.04\%} in mIoU on both tasks.
Excluding DM results in failure on RSRES, as the model loses structural priors necessary for segment object regions.
Exp.(\textit{w/o} Evolve) highlight the benefit of attention evolve module, increasing the Pr@0.5 and mIoU of \textbf{7.52\%} and \textbf{6.27\%} on RSREC.
The improved performance of O-F-E over O-E-F indicates that embedding the DM's self-attention into the VLM's cross-attention better preserves structural and shape information of objects, which facilitates more accurate region aggregation via DFS and suppresses noisy activations.

\noindent \textbf{Effect of self-attention resolutions.}
As shown in Tab.~\ref{tab:ablation2}, we investigate the impact of varying resolutions of self-attention maps.
We find that using multiple resolutions [32, 64] leads to the best performance, suggesting that multi-scale self-attention maps not only preserve high-resolution details but also effectively capture contextual semantics.

\begin{table}[!t]
\centering
\setlength{\tabcolsep}{1.8mm}
{\fontsize{9pt}{9.5pt}\selectfont
\begin{tabular}{c|cc|cc}
    \specialrule{.1em}{.3em}{.3em} 
    \multicolumn{1}{c|}{}                        & \multicolumn{2}{c|}{RSREC} & \multicolumn{2}{c}{RSRES}\\
    \cmidrule(r){2-5}       
    \multicolumn{1}{l|}{Resolution}  & Pr@0.5 & mIoU & Pr@0.5 & mIoU\\
    \specialrule{.1em}{.3em}{.3em}
    16              & 26.16 & 28.39 & 12.01 & 19.98 \\
    32              & 29.84 & 31.97 & 12.53 & 21.11   \\
    64              & 27.70 & 30.76 & 10.55 & 20.13 \\
    16, 32          & 28.23 & 29.86 & 12.70 & 20.41 \\
    32, 64          & \textbf{30.15} & \textbf{32.92} & \textbf{12.84} & \textbf{21.85}   \\
    16, 32, 64      & 27.86 & 30.51 & 11.98 & 20.36 \\
    \specialrule{.1em}{.3em}{.3em} 
\end{tabular}}
\caption{Effect of varying self-attention map resolutions.}
\label{tab:ablation2}
\end{table}

\section{Conclusion}
\label{sec5}
In this paper, we introduce RSVG-ZeroOV, a zero-shot framework that integrates a VLM and a DM via attention maps to tackle the open-vocabulary remote sensing visual grounding.
Without additional training, we design a novel overview-focus-evolve strategy to progressively enhance attention toward referred objects.
Extensive experiments on existing benchmarks demonstrate that RSVG-ZeroOV achieves SOTA performance, outperforming existing weakly-supervised and zero-shot methods.
To the best of our knowledge, RSVG-ZeroOV is the first zero-shot framework for remote sensing visual grounding.
We hope this work can potentially inspire researchers to address open-vocabulary remote sensing tasks from a fresh perspective.

\section{Acknowledgments}
This work was supported in part by the National Science and Technology Major Project under Grant 2022ZD0117103, in part by the National Natural Science Foundation of China under Grants 62577041, in part by the Outstanding Youth Science Foundation of Shaanxi Province under Grant 2025JC-JCQN-083, in part by the Key Research and Development Program of Shaanxi Province under Grants 2025CY-YBXM-047 and 2024GX-YBXM-140.

\bibliography{aaai2026}

@article{ding2024visual,
  title={Visual Selection and Multistage Reasoning for RSVG},
  author={Ding, Yueli and Xu, Haojie and Wang, Di and Li, Ke and Tian, Yumin},
  journal={IEEE Geoscience and Remote Sensing Letters},
  volume={21},
  pages={1--5},
  year={2024},
  publisher={IEEE}
}

@article{li2024show,
  title={Show me what and where has changed? question answering and grounding for remote sensing change detection},
  author={Li, Ke and Dong, Fuyu and Wang, Di and Li, Shaofeng and Wang, Quan and Gao, Xinbo and Chua, Tat-Seng},
  journal={arXiv preprint arXiv:2410.23828},
  year={2024}
}

@article{li2025segearth_r1,
  title={Segearth-r1: Geospatial pixel reasoning via large language model},
  author={Li, Kaiyu and Xin, Zepeng and Pang, Li and Pang, Chao and Deng, Yupeng and Yao, Jing and Xia, Guisong and Meng, Deyu and Wang, Zhi and Cao, Xiangyong},
  journal={arXiv preprint arXiv:2504.09644},
  year={2025}
}

@inproceedings{li2025segearth_ov,
  title={Segearth-ov: Towards training-free open-vocabulary segmentation for remote sensing images},
  author={Li, Kaiyu and Liu, Ruixun and Cao, Xiangyong and Bai, Xueru and Zhou, Feng and Meng, Deyu and Wang, Zhi},
  booktitle={Proceedings of the Computer Vision and Pattern Recognition Conference},
  pages={10545--10556},
  year={2025}
}

@inproceedings{sun2022visual,
  title={Visual grounding in remote sensing images},
  author={Sun, Yuxi and Feng, Shanshan and Li, Xutao and Ye, Yunming and Kang, Jian and Huang, Xu},
  booktitle={Proceedings of the 30th ACM International Conference on Multimedia},
  pages={404--412},
  year={2022}
}

@article{zhan2023rsvg,
  title={Rsvg: Exploring data and models for visual grounding on remote sensing data},
  author={Zhan, Yang and Xiong, Zhitong and Yuan, Yuan},
  journal={IEEE Transactions on Geoscience and Remote Sensing},
  volume={61},
  pages={1--13},
  year={2023},
  publisher={IEEE}
}

@article{li2020object,
  title={Object detection in optical remote sensing images: A survey and a new benchmark},
  author={Li, Ke and Wan, Gang and Cheng, Gong and Meng, Liqiu and Han, Junwei},
  journal={ISPRS journal of photogrammetry and remote sensing},
  volume={159},
  pages={296--307},
  year={2020},
  publisher={Elsevier}
}

@article{li2024language,
  title={Language-Guided Progressive Attention for Visual Grounding in Remote Sensing Images},
  author={Li, Ke and Wang, Di and Xu, Haojie and Zhong, Haodi and Wang, Cong},
  journal={IEEE Transactions on Geoscience and Remote Sensing},
  year={2024},
  publisher={IEEE}
}

@inproceedings{liu2024rotated,
  title={Rotated multi-scale interaction network for referring remote sensing image segmentation},
  author={Liu, Sihan and Ma, Yiwei and Zhang, Xiaoqing and Wang, Haowei and Ji, Jiayi and Sun, Xiaoshuai and Ji, Rongrong},
  booktitle={Proceedings of the IEEE/CVF Conference on Computer Vision and Pattern Recognition},
  pages={26658--26668},
  year={2024}
}

@inproceedings{kirillov2023segment,
  title={Segment anything},
  author={Kirillov, Alexander and Mintun, Eric and Ravi, Nikhila and Mao, Hanzi and Rolland, Chloe and Gustafson, Laura and Xiao, Tete and Whitehead, Spencer and Berg, Alexander C and Lo, Wan-Yen and others},
  booktitle={Proceedings of the IEEE/CVF International Conference on Computer Vision},
  pages={4015--4026},
  year={2023}
}

@inproceedings{radford2021learning,
  title={Learning transferable visual models from natural language supervision},
  author={Radford, Alec and Kim, Jong Wook and Hallacy, Chris and Ramesh, Aditya and Goh, Gabriel and Agarwal, Sandhini and Sastry, Girish and Askell, Amanda and Mishkin, Pamela and Clark, Jack and others},
  booktitle={International Conference on Machine Learning},
  pages={8748--8763},
  year={2021},
  organization={PMLR}
}

@inproceedings{huang2025zori,
  title={ZoRI: Towards discriminative zero-shot remote sensing instance segmentation},
  author={Huang, Shiqi and He, Shuting and Wen, Bihan},
  booktitle={Proceedings of the AAAI Conference on Artificial Intelligence},
  volume={39},
  number={4},
  pages={3724--3732},
  year={2025}
}

@inproceedings{shan2025ros,
  title={Ros-sam: High-quality interactive segmentation for remote sensing moving object},
  author={Shan, Zhe and Liu, Yang and Zhou, Lei and Yan, Cheng and Wang, Heng and Xie, Xia},
  booktitle={Proceedings of the Computer Vision and Pattern Recognition Conference},
  pages={3625--3635},
  year={2025}
}

@article{liu2023visual,
  title={Visual instruction tuning},
  author={Liu, Haotian and Li, Chunyuan and Wu, Qingyang and Lee, Yong Jae},
  journal={Advances in neural information processing systems},
  volume={36},
  pages={34892--34916},
  year={2023}
}

@article{chen2024rsprompter,
  title={RSPrompter: Learning to prompt for remote sensing instance segmentation based on visual foundation model},
  author={Chen, Keyan and Liu, Chenyang and Chen, Hao and Zhang, Haotian and Li, Wenyuan and Zou, Zhengxia and Shi, Zhenwei},
  journal={IEEE Transactions on Geoscience and Remote Sensing},
  volume={62},
  pages={1--17},
  year={2024},
  publisher={IEEE}
}

@inproceedings{liu2024grounding,
  title={Grounding dino: Marrying dino with grounded pre-training for open-set object detection},
  author={Liu, Shilong and Zeng, Zhaoyang and Ren, Tianhe and Li, Feng and Zhang, Hao and Yang, Jie and Jiang, Qing and Li, Chunyuan and Yang, Jianwei and Su, Hang and others},
  booktitle={European Conference on Computer Vision},
  pages={38--55},
  year={2024},
  organization={Springer}
}

@inproceedings{pan2025locate,
  title={Locate anything on earth: Advancing open-vocabulary object detection for remote sensing community},
  author={Pan, Jiancheng and Liu, Yanxing and Fu, Yuqian and Ma, Muyuan and Li, Jiahao and Paudel, Danda Pani and Van Gool, Luc and Huang, Xiaomeng},
  booktitle={Proceedings of the AAAI Conference on Artificial Intelligence},
  volume={39},
  number={6},
  pages={6281--6289},
  year={2025}
}

@article{lan2024language,
  title={Language query-based transformer with multiscale cross-modal alignment for visual grounding on remote sensing images},
  author={Lan, Meng and Rong, Fu and Jiao, Hongzan and Gao, Zhi and Zhang, Lefei},
  journal={IEEE Transactions on Geoscience and Remote Sensing},
  volume={62},
  pages={1--13},
  year={2024},
  publisher={IEEE}
}

@article{dong2024cross,
  title={Cross-modal bidirectional interaction model for referring remote sensing image segmentation},
  author={Dong, Zhe and Sun, Yuzhe and Liu, Tianzhu and Zuo, Wangmeng and Gu, Yanfeng},
  journal={arXiv preprint arXiv:2410.08613},
  year={2024}
}

@inproceedings{yao2025remotesam,
  title={RemoteSAM: Towards Segment Anything for Earth Observation},
  author={Yao, Liang and Liu, Fan and Chen, Delong and Zhang, Chuanyi and Wang, Yijun and Chen, Ziyun and Xu, Wei and Di, Shimin and Zheng, Yuhui},
  booktitle={Proceedings of the 33th ACM International Conference on Multimedia},
  year={2025}
}

@article{yuan2023rrsis,
  title={Rrsis: Referring remote sensing image segmentation},
  author={Yuan, Zhenghang and Mou, Lichao and Hua, Yuansheng and Zhu, Xiao Xiang},
  journal={arXiv preprint arXiv:2306.08625},
  year={2023}
}

@article{wang2024cogvlm,
  title={Cogvlm: Visual expert for pretrained language models},
  author={Wang, Weihan and Lv, Qingsong and Yu, Wenmeng and Hong, Wenyi and Qi, Ji and Wang, Yan and Ji, Junhui and Yang, Zhuoyi and Zhao, Lei and XiXuan, Song and others},
  journal={Advances in Neural Information Processing Systems},
  volume={37},
  pages={121475--121499},
  year={2024}
}

@inproceedings{chen2024internvl,
  title={Internvl: Scaling up vision foundation models and aligning for generic visual-linguistic tasks},
  author={Chen, Zhe and Wu, Jiannan and Wang, Wenhai and Su, Weijie and Chen, Guo and Xing, Sen and Zhong, Muyan and Zhang, Qinglong and Zhu, Xizhou and Lu, Lewei and others},
  booktitle={Proceedings of the IEEE/CVF conference on computer vision and pattern recognition},
  pages={24185--24198},
  year={2024}
}

@article{wang2024qwen2,
  title={Qwen2-vl: Enhancing vision-language model's perception of the world at any resolution},
  author={Wang, Peng and Bai, Shuai and Tan, Sinan and Wang, Shijie and Fan, Zhihao and Bai, Jinze and Chen, Keqin and Liu, Xuejing and Wang, Jialin and Ge, Wenbin and others},
  journal={arXiv preprint arXiv:2409.12191},
  year={2024}
}

@article{chiang2023vicuna,
  title={Vicuna: An open-source chatbot impressing gpt-4 with 90\%* chatgpt quality},
  author={Chiang, Wei-Lin and Li, Zhuohan and Lin, Ziqing and Sheng, Ying and Wu, Zhanghao and Zhang, Hao and Zheng, Lianmin and Zhuang, Siyuan and Zhuang, Yonghao and Gonzalez, Joseph E and others},
  journal={See https://vicuna. lmsys. org (accessed 14 April 2023)},
  volume={2},
  number={3},
  pages={6},
  year={2023}
}

@inproceedings{rombach2022high,
  title={High-resolution image synthesis with latent diffusion models},
  author={Rombach, Robin and Blattmann, Andreas and Lorenz, Dominik and Esser, Patrick and Ommer, Bj{\"o}rn},
  booktitle={Proceedings of the IEEE/CVF conference on computer vision and pattern recognition},
  pages={10684--10695},
  year={2022}
}

@inproceedings{pnvr2023ld,
  title={Ld-znet: A latent diffusion approach for text-based image segmentation},
  author={Pnvr, Koutilya and Singh, Bharat and Ghosh, Pallabi and Siddiquie, Behjat and Jacobs, David},
  booktitle={Proceedings of the IEEE/CVF international conference on computer vision},
  pages={4157--4168},
  year={2023}
}

@inproceedings{xu2023open,
  title={Open-vocabulary panoptic segmentation with text-to-image diffusion models},
  author={Xu, Jiarui and Liu, Sifei and Vahdat, Arash and Byeon, Wonmin and Wang, Xiaolong and De Mello, Shalini},
  booktitle={Proceedings of the IEEE/CVF conference on computer vision and pattern recognition},
  pages={2955--2966},
  year={2023}
}

@article{zheng2025instructsam,
  title={InstructSAM: A Training-Free Framework for Instruction-Oriented Remote Sensing Object Recognition},
  author={Zheng, Yijie and Wu, Weijie and Li, Qingyun and Wang, Xuehui and Zhou, Xu and Ren, Aiai and Shen, Jun and Zhao, Long and Li, Guoqing and Yang, Xue},
  journal={arXiv preprint arXiv:2505.15818},
  year={2025}
}

@article{liu2024remoteclip,
  title={Remoteclip: A vision language foundation model for remote sensing},
  author={Liu, Fan and Chen, Delong and Guan, Zhangqingyun and Zhou, Xiaocong and Zhu, Jiale and Ye, Qiaolin and Fu, Liyong and Zhou, Jun},
  journal={IEEE Transactions on Geoscience and Remote Sensing},
  volume={62},
  pages={1--16},
  year={2024},
  publisher={IEEE}
}

@inproceedings{ye2025towards,
  title={Towards open-vocabulary remote sensing image semantic segmentation},
  author={Ye, Chengyang and Zhuge, Yunzhi and Zhang, Pingping},
  booktitle={Proceedings of the AAAI Conference on Artificial Intelligence},
  volume={39},
  number={9},
  pages={9436--9444},
  year={2025}
}

@inproceedings{liu2024improved,
  title={Improved baselines with visual instruction tuning},
  author={Liu, Haotian and Li, Chunyuan and Li, Yuheng and Lee, Yong Jae},
  booktitle={Proceedings of the IEEE/CVF conference on computer vision and pattern recognition},
  pages={26296--26306},
  year={2024}
}

@inproceedings{kuckreja2024geochat,
  title={Geochat: Grounded large vision-language model for remote sensing},
  author={Kuckreja, Kartik and Danish, Muhammad Sohail and Naseer, Muzammal and Das, Abhijit and Khan, Salman and Khan, Fahad Shahbaz},
  booktitle={Proceedings of the IEEE/CVF Conference on Computer Vision and Pattern Recognition},
  pages={27831--27840},
  year={2024}
}

@article{Qwen2.5-VL,
  title={Qwen2.5-VL Technical Report},
  author={Bai, Shuai and Chen, Keqin and Liu, Xuejing and Wang, Jialin and Ge, Wenbin and Song, Sibo and Dang, Kai and Wang, Peng and Wang, Shijie and Tang, Jun and Zhong, Humen and Zhu, Yuanzhi and Yang, Mingkun and Li, Zhaohai and Wan, Jianqiang and Wang, Pengfei and Ding, Wei and Fu, Zheren and Xu, Yiheng and Ye, Jiabo and Zhang, Xi and Xie, Tianbao and Cheng, Zesen and Zhang, Hang and Yang, Zhibo and Xu, Haiyang and Lin, Junyang},
  journal={arXiv preprint arXiv:2502.13923},
  year={2025}
}

\end{document}